# Multi-scale multi-modal micro-expression recognition algorithm based on transformer

Fengping Wang, Jie Li[*], Chun Qi, Lin Wang, Pan Wang

*School of Information and Communications Engineering, Xi'an Jiaotong University, Xi'an, China*

**Abstract:** A micro-expression is a spontaneous unconscious facial muscle movement that can reveal the true emotions people attempt to hide. Although manual methods have made good progress and deep learning is gaining prominence. Due to the short duration of micro-expression and different scales of expressed in facial regions, existing algorithms cannot extract multi-modal multi-scale facial region features while taking into account contextual information to learn underlying features. Therefore, in order to solve the above problems, a multi-modal multi-scale algorithm based on transformer network is proposed in this paper, aiming to fully learn local multi-grained features of micro-expressions through two modal features of micro-expressions - motion features and texture features. To obtain local area features of the face at different scales, we learned patch features at different scales for both modalities, and then fused multi-layer multi-headed attention weights to obtain effective features by weighting the patch features, and combined cross-modal contrastive learning for model optimization. We conducted comprehensive experiments on three spontaneous datasets, and the results show the accuracy of the proposed algorithm in single measurement SMIC database is up to 78.73% and the F1 value on CASMEII of the combined database is up to 0.9071, which is at the leading level.

# 1 Introduction

Micro-expressions, as a kind of human emotion, can express people's inner real feelings. Therefore, the detection and recognition of micro-expressions have been widely used in clinical diagnosis, criminal investigation analysis, and national defense and security. The short duration of micro-expressions, no more than 1/2s, and occurring only in localized areas of the face [1,2], makes detecting and recognizing micro-expressions difficult.

Early micro-expression studies were based on manual features and there are two main categories: texture feature algorithms based on Local Binary Patterns-Three Orthogonal Planes (LBP-TOP) and motion feature algorithms based on optical flow (OF). In recent years, deep learning algorithms have been the main trend in micro-expression recognition research. Convolutional neural network (CNN)-based algorithms are used to extract spatiotemporal motion

______________

*Corresponding author.
E-mail: jielixjtu@xjtu.edu.cn.

features of micro-expressions, while also combining them with traditional features. In terms of feature extraction, since facial expressions are triggered by facial muscle unit movements, mining micro-features from local spatial regions is one of the ways to study. For example, LBP-TOP features with multi-scale activation patches [3,4,5], Main Directional Mean Optical Flow (MDMO) [6,7], Local region feature learning [8,9], and so on. However, although these efforts emphasize the task of local region feature extraction, there are still some drawbacks that need to be addressed. For instance, 1) multi-scale optical flow features cannot fully represent the changing features of expressions, or different modal features cannot effectively capture local features even by using fixed patches. 2)treating all key regions equally ignores the validity of the representation of local features and the connection relationship between local blocks and expressions. In order to extract local features of micro-expressions more rationally and to learn the relationship between local features, transformer has been introduced into the study of micro-expressions [10,11,12,13], and good results have been obtained. However, these algorithms will all utilize fixed-size patches and do not take into account local area features at different scales of micro-expressions. Therefore, to solve the above problem, we applied the multi-headed self-attention mechanism transformer to build our network framework. This network attempts to perform multi-scale patches on the inputs of two different modalities and learn meaningful facial features by transformer self-attention property. The results confirm that this method can learn the subtle features of micro-expressions with better performance than the current level. The main contributions of this paper are summarized as follows:

1: We proposed a multi-scale multi-modality transformer network-MSMMT for learning micro-expression features. To learn facial expression features at different scales, we first acquire images of different scales and then input them to the transformer for patch embedding to extract multi-scale features.

2: Considering the different degrees of the contribution of features to micro-expressions at different scales, the patches attention mechanism is proposed after multi-scale feature extraction. The core idea is to multiply the weights learned in the first N-1 layers of the Transformer and then weight the patch features in the last layer to obtain the most effective region features.

3: We apply unsupervised contrastive learning loss to make similar features closer and dissimilar features away in two multi-scale features, and use cross-entropy loss for joint features to optimize the network. The effectiveness of this algorithm is experimentally demonstrated.

The rest of the article is organized as follows: Section 2 presents the related work. Section 3 is the proposed method. The experimental results are given in Section 4. Finally, we conclude the paper in Section 5.

# 2 Related works

2.1 Micro-expression recognition

Traditional manual feature extraction algorithms based on local regions are the main tools in the preliminary stage of micro-expression research, and local blocking of images or sequences using different methods is the main way to extract local features of the face. Wang et al. [14] first extracted the fine texture motion features of the sparse part of the micro-expression using robust principal component analysis (RPCA), then extracted the local texture orientation features of the 16 ROIs using a local spatiotemporal algorithm. Sun et al. [15] proposed a spatiotemporal LBP-TOP descriptor for multi-scale patch fusion, which considers the active contribution of different regional areas of the face. To better capture the low-intensity image features corresponding to small local areas, Zong et al. [16] used a multi-scale spatial segmentation grid to segment video clips into multi-level local blocks to extract spatiotemporal descriptors. A kernelized group sparse learning (KGSL) model is then used to learn more efficient multi-level spatiotemporal descriptors. Liu et al. [17] proposed the MDMO, a normalized statistical feature based on the region of interest (ROI), and extracted optical flow statistical features for the 36 ROI regions of the face. Liu et al. [18] proposed sparse MDMO, whose core idea is to introduce classical graph regularized sparse encoding in the MDMO feature space. The article captures this sparsity with a new distance metric. Allaert et al. [19] proposed a new feature extraction algorithm-Local Motion Pattern (LMP)-which performs a local analysis of the motion distribution to separate consistent motion patterns from the noise. Since LMP extracts local features based on 25 ROI regions divided by the motion pattern of facial expressions, the method is applicable to handle all expressions that cause facial skin deformation. Liong et al. [20] proposed an optical strain (OS) weighted feature extraction method for subtle expression recognition of human faces. OS has better recognition of deformation results than OF and can better represent the motion characteristics of micro-expressions. In the article, the micro-expression sequences are locally blocked to extract local LBP-TOP features, and then the contribution values of different regions are weighted by optical strain. To reduce redundancy, Liong et al. [1] extracted double-weighted directional OF features for the Apex frames of micro-expression sequences. Similarly, the algorithm still uses OS for local region feature weighting.

Micro-expression deep learning algorithms have become more popular in these years. In addition to end-to-end learning using original images or sequences, many scholars also input expression feature maps into networks for learning research. And these algorithms utilize different ways to extract the local features of micro-expressions. Wang et al. [21] used optical flow to capture small changes in micro-expressions as they occur and then designed convolution kernels of different scales to extract local features of micro-expressions of different intensities. Liong et al. [22] also proposed a shallow three-stream 3D CNN (STSTNet) to extract discriminative high-level and detailed micro-expressions features. The network learns three optical flow features (i.e., OS, horizontal optical flow field, and vertical optical flow field) based on the onset and apex frames computed from each video. This also solves the problem of insufficient samples. Li et al. [23] proposed a three-stream CNN (TSCNN) to fuse temporal, spatial, and local region cues of micro-expression videos to learn micro-expression salient features. Xia et al. [24] proposed to capture the spatiotemporal deformation features of expression sequences in a deep recurrent convolutional network (STRCN) based on the local area features. To obtain useful local area features, the temporal differences of video frames on the whole database were first accumulated to calculate a differential heat map based on the entire database, and the local perceptual area of micro-expressions was obtained after the thresholding process. Kumar et al. [25] proposed an end-to-end marker point-assisted dual-stream graph-attentive convolutional network for micro-expression feature recognition, where one stream uses the coordinate positions of facial marker points to construct graph nodes and edges to learn facial muscle movements, and one stream uses the local area optical flow features of marker points to learn facial features. Similarly, graph convolutional network-based methods are also available in [26]. Su et al. [27] proposed a micro-expression recognition method based on key facial components guidance (KF-MER). The idea is to divide the face into semantic regions, obtain division probability maps, learn the relationship between parts, and then use shallow residual networks for micro-expression learning. Li et al. [28] fed optical flow images into a multi-scale joint network for feature extraction and classification. The proposed joint feature module integrates features at different levels and facilitates capturing micro-expression features of various magnitudes. In addition, some works have used attention mechanisms to obtain local ROI features of faces. Yang et al. [29] proposed that MERTA integrates three types of attentional mechanisms - general attention, motivation attention, and channel attention - to extract landmark areas, motion regions, and expression-related semantic features, respectively. Zhang et al. [30] designed spatial attention, channel attention, and self-attentive ternary attention modules in a network to learn meaningful optical flow features. Wang et al. [31] proposed a micro-attention mechanism in concert with the residual network. Micro-attention enables the neural network to learn to attend to regions of interest of faces covering different action units. The self-output of each residual block at different scales is used to compute the attention map.

2.2 Vision Transformer

Transformer is a very successful model proposed in NLP in 2017 and subsequently used in computer vision with notable success. Visual transformer (ViT) can transform images into sequences by dividing them into sub-images and sorting them consistently, so that spatial correlations can be learned like temporal features, and then image classification can be performed [32]. The most significant difference between Transformer and CNN is that it uses a self-attention mechanism. It allows each token to represent contextual information in the group it belongs to rather than representing a single meaning. Since self-attention models the relationship of patches in an image, it is more expressive. Recently, researchers have used the transformer for various computer vision tasks and obtained remarkable results, and micro-expression algorithms related to the transformer have also been studied. Hong et al. [10] proposed a long and short-term correlation based on a spatiotemporal transformer for micro-expression recognition. The architecture includes a spatial encoder for learning spatial patterns, a temporal aggregator for temporal dimensional analysis, and a classification head. Zhang et al. [11] proposed a late-fusion-based transformer for expression recognition algorithm for video, where OF and grayscale sequences are fused after learning by the transformer, respectively. Both methods obtained good results. Li et al. [12] proposed two branching modules for micro-expression recognition. The transformer-based module was used for position calibration, and the continuous attention-based module was used for learning motion features. Zhu et al. [13] proposed a spatio-temporal feature learning method based on sparse transformation to obtain effective features of facial micro-expressions. Its core is to extract strongly correlated spatiotemporal features of micro-expression categories using spatial and temporal attention mechanisms, reducing influence of the irrelevant features.

## 3 The proposed approach

We proposed a multi-scale transformer-based feature extraction algorithm to learn more discriminative local region features of micro-expressions. The algorithm takes the dynamic imaging and the optical flow-optical strain images as inputs. To select important local region features, inspired by the RAMS-Trans[33] algorithm, the attention weights learned in all previous layers except the last layer of ViT are processed. Then the features are weighted and input to the last encoder layer for learning and classification. The overall process is shown in Figure 1.

3.1 dynamic imaging

Micro-expressions are fast and fleeting, appearing in only a few frames. These temporary offsets can be obtained from the video using dynamic imaging methods [34,35,36]. A dynamic imaging is a standard RGB image that preserves the spatial and temporal information of an entire video sequence into a single image [37]. To construct a dynamic imaging, a sorting function is applied to the video frames. The RGB feature vector of each frame $F_T$ is expressed as $\sigma(F_T)$.

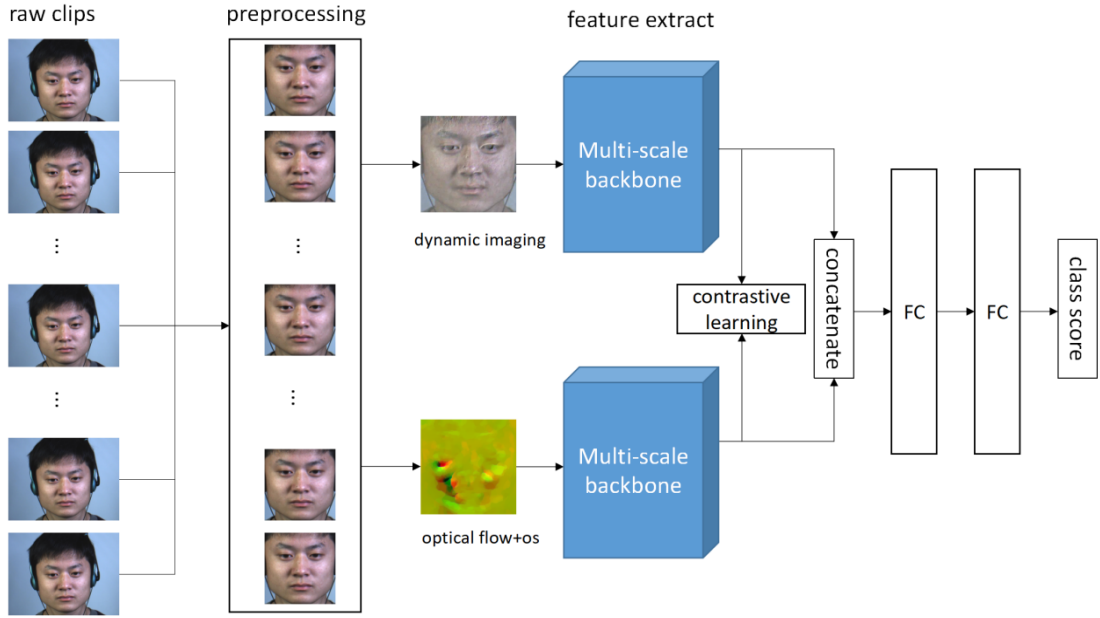

Figure 1 Overall framework of the algorithm

The temporal average of the available feature vectors ($\varphi_t$) is calculated using equation (1).

$$\varphi_t = \frac{1}{t}\sum_{T=1}^{t}\sigma(F_T) \quad (1)$$

The score related to time t is then calculated by the sorting function, as in Equation (2).

$$\psi(t|d) = \langle d, \phi_t \rangle \quad (2)$$

Where $d \in R^d$ represents a vector to calculate the fraction of frames in the video. Higher ranks are assigned to frames of time l, i.e. $(l>t) \Rightarrow \psi(l|d) > \psi(t|d)$ [35]. Finally, RankSVM is used to calculate out d, as in Equation (3) and Equation (4).

$$d^* = \eta(F_1, F_2, \cdots, F_T; \sigma) = \arg\min(E(D)) \quad (3)$$

$$E(D) = \frac{d}{2}\|d\|^2 + \frac{2}{T(T-1)} \times \sum_{l>t} \max\{0, 1-\psi(l|d)+\psi(t|d)\} \quad (4)$$

Equation (3) defines a function $\eta(F_1, F_2, \cdots, F_T; \sigma)$ that converts the video frames into a single vector d*. Thus, d* combines the details of all frames and is often used as a video descriptor. Equation (4) is the solution of two key functions: a quadratic regularize implemented in SVMs, and a hinge-loss soft-counting function that indicates how many pairs $(l>t)$ are misaligned by the rank function [36].

The dynamic imaging of micro-expressions is shown in Figure 2. The figure shows that the generated dynamic image successfully preserves the consistent and inconsistent information of different categories of expressions within a single frame.

### 3.2 OF component

The concept of OF refers to the movement of target pixels in an image due to the movement of objects in the image or the movement of the camera in two consecutive frames. It encodes the motion of the object in vector notation, representing the flow direction and intensity of each image pixel. There have been many feature extraction algorithms on optical flow in existing micro-expression studies, and it has been confirmed that OF can extract subtle facial change features of micro-expressions [1,38]. The mathematical definition of optical flow is as follows.

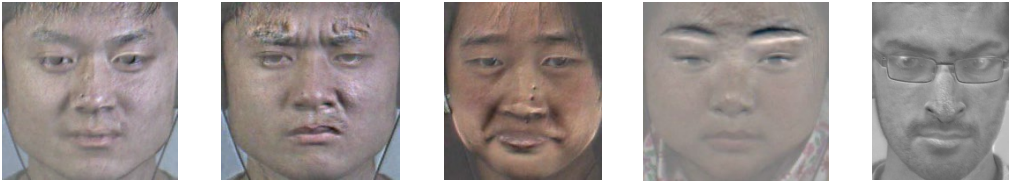

Figure 2 Dynamic images of different micro-expression expressions, from left to right: happiness, disgust, repression, surprise, anger, sadness.

$$\nabla I \cdot \vec{p} + I_t = 0 \tag{5}$$

where $I(x,y,t)$ denotes the pixel intensity of the space $(x,y)$ at time t. $\nabla I = (I_x, I_y)$ refers to the spatial gradient and $I_t$ denotes the temporal gradient representing the intensity function. $\vec{p}$ is the horizontal and vertical components of the optical flow, denoted as follows.

$$\vec{p} = \left[ p = \frac{dx}{dt}, q = \frac{dy}{dt} \right]^T \tag{6}$$

which $dx$, $dy$ refers to the change in the two-dimensional position and $d_t$ represents the change in time. This constraint on luminance constancy assumes that the pixel intensities of the two images are constant over time. We adopt TV-L1 [39] for optical flow approximation because it has two main advantages: better noise robustness and the ability to preserve the discontinuity of the flow.

To obtain essential features of optical flow features in different directions, we compute optical strain(OS) for feature learning. Each video can be represented by a single OS image showing facial deformation [40]. The two-dimensional displacement vector of the moving object can be expressed as $u = [u,v]^T$. Assuming that the moving object is in a small motion, the strain magnitude can be defined as Equation (7).

$$\varepsilon = \frac{1}{2}\left[ \nabla u + (\nabla u)^T \right] \tag{7}$$

It can be further expanded as

$$\varepsilon = \begin{bmatrix} \varepsilon_{xx} = \frac{\partial u}{\partial x} & \varepsilon_{xy} = \frac{1}{2}\left(\frac{\partial u}{\partial y} + \frac{\partial v}{\partial x}\right) \\ \varepsilon_{yx} = \frac{1}{2}\left(\frac{\partial v}{\partial x} + \frac{\partial u}{\partial y}\right) & \varepsilon_{yy} = \frac{\partial v}{\partial y} \end{bmatrix} \tag{8}$$

The diagonal strain component $(\varepsilon_{xx}, \varepsilon_{yy})$ is the normal strain component, which $(\varepsilon_{xy}, \varepsilon_{yx})$ is shear strain component. Specifically, normal strain measures the change in length along a particular direction. And shear strain measures the change in two angles.

The optical strain for each pixel can be obtained by calculating the sum of the squares of the normal and shear strain components, as shown in equation (9).

$$|\varepsilon_{x,y}| = \sqrt{\varepsilon_{xx}^2 + \varepsilon_{yy}^2 + \varepsilon_{xy}^2 + \varepsilon_{yx}^2} \\ = \sqrt{\frac{\partial u^2}{\partial x} + \frac{\partial v^2}{\partial x} + \frac{1}{2}\left(\frac{\partial u}{\partial x} + \frac{\partial u}{\partial x}\right)^2} \tag{9}$$

To obtain the first- and second-order optical flow variation features, referring to [22,41], we use the horizontal optical flow, vertical optical flow, and optical strain as three channels of the image for feature learning, as shown in Figure 3.

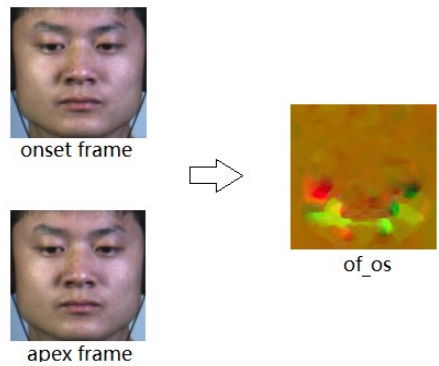

Figure 3 Optical flow-optical strain image

### 3.3 Multi-scale multi-modal transformer

To obtain multi-scale features of images, this paper converts images into multiple scales, then learns the spatial relationship between different patches through the network self-attention mechanism to get local discriminative features of the same modal at different scales and improve the recognition ability of the model. The multi-scale module and the specific patch feature weighted module are shown in figure 4 and figure 5.

*3.3.1 Multi-scale patch*

The original ViT model has a fixed size on the patch, which tends to ignore finer-grained micro-expression features, despite the results achieved. To acquire images at different scales, as shown in figure 4, the image is converted to the specified size $(m,n)$ for patches at the first scale, which can be divided into $(m/16)\times(m/16)$ patches. At the second scale, the image is reduced to $(m/2,n/2)$ size, still patched embedding according to the pixel size of $16\times16$, and at the third scale, the image is reduced to $(m/4,n/4)$ size and patched embedding in the same way. Since ViT is used to learn the spatial relationships of different patch regions, we do not consider $1\times1$ patch.

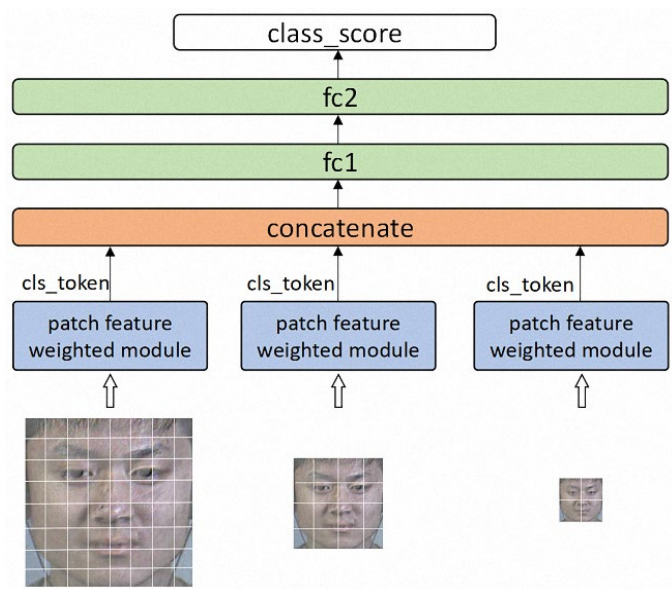

Figure 4 Multi-scale feature selection module

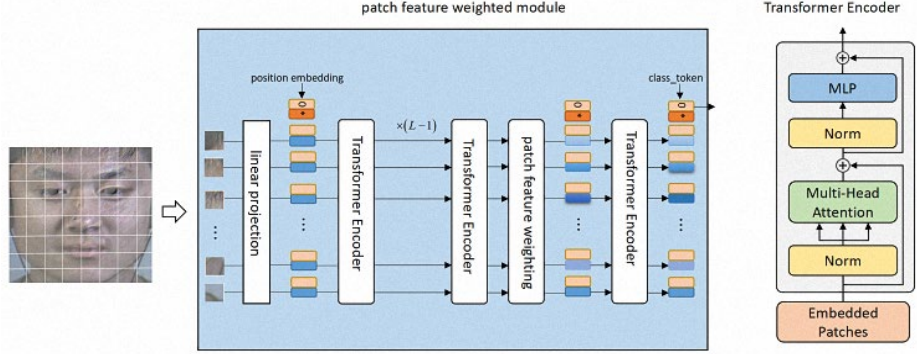

Figure 5 patch feature weighting module

We proposed a new method of patch selection in which the intensity of attention can be intuitively used as an indicator to characterize the importance of symbols. As shown in Figure 5, in this module, we integrate all the original attention weights of the first $L-1$ layers of the transformer into an attention graph to guide the network to efficiently and accurately select the distinguished image patches and compute their relationships. The embedding lacks token identifiability, and the original attention weights do not necessarily correspond to the relative importance of the input token, especially for the higher levels of the model [42]. To effectively utilize the attention mechanism of ViT, we improved the input of the last layer of the network. As shown in Figure 6, the weights of the first $L-1$ layers of the network are first extracted.

$$w_l = \left[w_l^1, w_l^2, \cdots, w_L^H\right], l \in 1,2,\cdots,L-1 \tag{10}$$

$$w_l^i = \left[w_l^1, w_l^2, \cdots, w_l^D\right], i \in 1,2,\cdots,H \tag{11}$$

where $H$ is the number of heads, $L$ is the number of layers, and $D$ is the embedded feature dimension.

We consider that the critical attention features are gradually accumulated and amplified in each layer, so we first regularized the weights of multiple attentions, that is, we sum up the multiple attention values of each layer weight and take the mean value, and then regularize them in each embedding dimension (Regularization is mainly used to avoid the generation of over-fitting and reduce network errors):

$$G_l = \frac{\frac{1}{H}\sum_{h=1}^{H} w_l^h}{\frac{1}{D}\sum_{d=1}^{D}\left(\frac{1}{H}\sum_{h=1}^{H} w_l^h\right)} \quad (12)$$

The attention weights of all previous layers are then integrated and the matrix multiplication operation is performed recursively on the modified attention weights of all layers as follows:

$$G = \prod_{L=1}^{L-1}(G_l) \quad (13)$$

Since the elements in G are the attention-related values of each patch with respect to other patches, to obtain the weight value of each patch, we average the results by row and normalize them:

$$\bar{G} = mean(G,0)/\max(mean(G,0)) \quad (14)$$

Finally, the output features of the L-1th layer are multiplied with the adjusted attention weights to obtain the weighted patch features, which are then concatenated with the cls_token and input to the last Transformer Layer.

$$input_L = \left[ cls\_token_{L-1}, \bar{G} Z_{L-1} \right] \quad (15)$$

The output of the last layer is shown in equation (16):

$$Z_L = FFN\left(LN\left(MSM\left(LN(input_L)\right)\right) + input_L\right) \quad (16)$$

Since facial expressions occur in some local regions of the face, other regional features contribute less to the expressions, but some features will be lost if the regions with lower weights are dropped, so these features are still kept. In this way, we not only keep the global information, but focus on the nuances of micro-expressions in the last Transformer Layer.

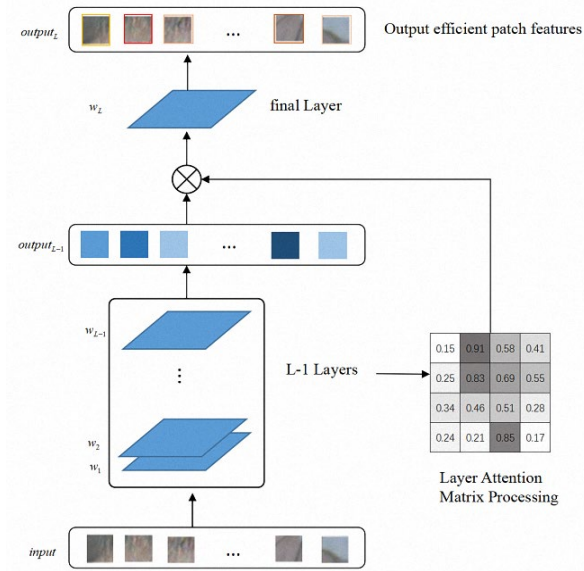

Figure 6 patch feature weighting

*3.3.3 Multi-feature fusion*

Since optical flow and dynamic imaging have different representations of micro-expression and have complementary roles, therefore, to capture the key features of different modalities in multi-scale mode, we concatenate the cls_token of the features learned at different scales as the final features of individual modal. The two features are then concatenated into two fully connected layers for expression classification, as shown in equations (17) and (18). To avoid over-fitting, we add dropout and ReLU after the first fully connected layer.

$$class\_score = FC_2\left(FC_1\left(Concat(dy\_imaging_{final}, flow\_os_{final})\right)\right) \quad (17)$$

$$F_{final} = Concat(cls\_token_1, cls\_token_2, cls\_token_3), F \in (dy\_imaging, flow\_os) \quad (18)$$

*3.3.4 Network optimization*

Contrast learning generates anchor samples, positive and negative samples from an unlabeled database and learns the similarity of the samples. The common features of the samples are extracted by encoding the positive samples similarly and encoding the negative samples differently through an encoder. Since the dynamic and optical flow maps used in this paper belong to two different modalities, the similarity of the two modal features on the same category is maximized and the similarity of different categories is minimized by using contrast learning [43].

The micro-expression database i consists of two modalities, the sample set $dy_i$ and $flow\_os_i$, and positive and negative sample pairs are constructed according to whether the two modalities come from the same micro-expression sample. For the positive sample pair, fixed $dy_i$ and the loss function is:

$$L_{i,dy,flow\_os} = -\log \frac{\exp(s(dy_i, flow\_os_i)/\tau)}{\sum_{k=1,k\neq i}^{B}\exp(s(dy_i, flow\_os_i)/\tau) + \sum_{k=1}^{B}\exp(s(dy_i, flow\_os_k)/\tau)} \quad (19)$$

where $B$ is the small batch size, $\tau$ is the temperature coefficient, and $s(.)$ denotes the cosine similarity function.

Fixed $flow\_os_i$, the loss can be obtained, and the comparative loss function $L_{con}$ for the two modes is then denoted as:

$$L_{con} = \frac{1}{2B}\sum_{i=1}^{B}[L_{i,dy,flow\_os} + L_{i,flow\_os,dy}] \quad (20)$$

The total loss of the network is:

$$L_{loss} = (1-\alpha)*L_{con} + \alpha*L_{cross\_entroy} \quad (21)$$

# 4 Experiments

4.1 Datasets

Three spontaneous micro-expression datasets are involved in the experimental validation of the algorithm in this paper, i.e., SMIC, CASMEII, SAMM. To be able to compare fairly with other state-of-the-art algorithms, we not only test the three datasets individually, but also refer to the combined data set proposed in MEG2019 [44] to measure the algorithm in this paper.

*A. SMIC*

The SMIC database [45] is the first database containing spontaneous micro-expressions obtained through an emotion elicitation experiment. The experimental test data is an HS subset with a frame rate of 100 frames and a resolution of $640 \times 480$. The data set includes a total of 164 micro-expression clips from 16 participants and does not contain action unit labels, and the data were labeled by two coders into three categories of expression clips based on participants' statements about their subjective emotions while watching the video: positive (51), negative (70), and surprised (43).

*B. CASMEII*

CASME II [46] is an improved version of the CASME database with a higher frame rate (200pfs) and a resolution of $640 \times 480$. The CASME II database contains 255 micro-expression clips from 26 subjects. Each micro-expression clip is classified by the coders according to the facial AU, the subject's self-reports and the video content, and the database contains seven categories, namely disgust(60), happiness(32), repression(27), surprise(25), sadness(7),fear(2) and others(102). Because there are few samples of sadness and fear, only the first five classes of data are selected in single data set experiment. In addition, the database provides the onset frame, apex frame, offset frame and AU labels of the micro-expression.

*C. SAMM*

SAMM [47] contains a total of 159 micro-expression sequences from 32 participants from 13 ethnic groups, a database with a high resolution of $2040 \times 1088$ and a frame rate of 200. The coders objectively classified expressions into seven basic expression categories based on FACS: anger (57), happiness (26), surprise (15), contempt (12), disgust (9), fear (8), sadness (6), and other (26). Similarly, in the single database test, the data participating in the experimental analysis only contains the five categories with the largest number of categories. SAMM also provides onset frames, apex frames, offset frames, and AU labels of expression clips.

The combined database included 442 micro-expression samples from 68 participants from three datasets. Of these, 164 samples were from SMIC, 145 were from CASMEII, and 133 were from SAMM. These samples were reclassified into three categories: positive, negative, and surprise [44]. Since the transformer network needs enough data for network learning, we enhance the data by rotating [-10,10], flipping horizontally, scale change, etc. The SMIC dataset does not provide the onset and apex frames of expressions, so we take the intermediate frames as the apex frames for feature extraction.

4.2 Evaluation metrics

Due to the imbalanced distribution of the number of categories in the database, we used three metrics to reduce bias: accuracy (Acc), unweighted average recall (UAR), and unweighted F1 score (UF1). The unweighted F1 score (UF1) is obtained by summing the F1 value of each class and then calculating the average value according to the number of classes. The unweighted average recall (UAR) is obtained by summing the accuracy of each class and then averaging over the number of classes.

$$Acc = \frac{\sum_{c=1}^{C} TP_c}{\sum_{c=1}^{C} N_c} \quad (22)$$

$$UAR = \frac{1}{C}\sum_{c=1}^{C} \frac{TP_c}{N_c} \quad (23)$$

$$UF1 = \frac{1}{C}\sum_{c=1}^{C} \frac{2TP_c}{2TP_c + FP_c + FN_c} \quad (24)$$

where C is the number of classes and $N_c$ is the number of samples per class. *TP*, *FN* and *FP* are true positives, false negatives and false positives, respectively.

In the experiments, we used the LOSO cross-validation method. All the micro-expression data of one subject were used for testing, and the samples of the other subjects were used for training until each subject completed the test as a test set.

4.3 Pre-processing

Since the SAMM database does not provide cropped data, we processed the three datasets in a unified manner to reduce interference terms in the micro-expression recognition. As shown in Figure 7, we used the Dlib library to locate 68 feature points of the face. Since the duration of micro-expressions is relatively short, some tight facial movements can be ignored. Therefore, we only compute the affine transformation for the first frame image with two inner eye angle coordinates, align the subsequent frames with the same transformation, and then perform facial cropping with the landmark points of the aligned face.

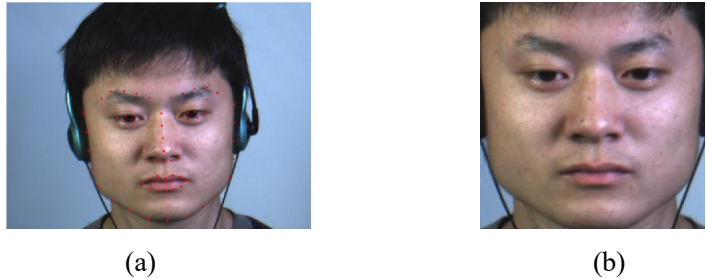

(a)          (b)

Figure 7 (a)68 landmark points on the face (b)cropped face

*4.3.1 Eulerian video magnification*

The Eulerian video amplification (EVM) technique [48] can be used to amplify subtle motion changes in video that are difficult to see with the naked eye. Motion amplification has been widely used in micro-expression recognition tasks due to the increased micro-motion amplitude, which is better for feature extraction and analysis [49,50]. In this paper, through experimental comparison, the optimal results can be obtained when the amplification factor is set as 10.

4.4 Implementation details

For multi-scale feature extraction process, we used the ViT-base with 12 layers of Encoder and hidden layers of size 768, with 12 heads. For initialization, we used the pre-trained official

ViT-B model on ImageNet[32]. In the training phase, we set the batch size to 16 and trained the network for 50 epochs. We initialized the learning rate to 5e-5 and weight decay to 0.05, updating the network weights using AdamW. AdamW is an optimization algorithm that uses adaptive learning rate gradient descent to make the network converge faster. All our experiments are windows systems and Nvidia GeForce RTX 1080Ti GPU.

4.5 Results and discussion

We compare the proposed approach with commonly used manual feature extraction methods and recently outstanding deep learning methods in single database experiment (SDE) and combined database experiment (CDE) settings based on widely used micro-expression databases: SMIC-HS, CASME II, SAMM.

*4.5.1 SDE*

To ensure consistency and fairness of comparison, SDE results for all methods were obtained under the same conditions, i.e., for the same number of samples, number of labels (classes), and using the same cross-validation method.

As can be easily seen from Table 1, among the experimental results under SDE settings, the

Table1 Micro-expression recognition results of the proposed MSMMT algorithm and the state-of-the-art method tested singularly on the three datasets

|  | SMIC-HS | | CASMEII | | SAMM | |
|---|---|---|---|---|---|---|
|  | Acc | F1 | Acc | F1 | Acc | F1 |
| LBP-TOP[10*] | 53.66 | 0.5384 | 46.46 | 0.4241 | - | - |
| MDMO[6](2016) | 61.5 | 0.406 | 51.0 | 0.418 | - | - |
| DiSTLBP-RIP[14](2017) | 63.41 | - | 64.78 | - | - | - |
| Bi-WOOF[1](2018) | 59.3 | 0.620 | 58.9 | 0.610 | 59.8 | 0.591 |
| DSSN[54](2019) | 63.41 | 0.6462 | 70.78 | 0.7297 | 57.35 | 0.4644 |
| STRCN[24](2019) | 53.1 | 0.514 | 56.0 | 0.542 | 54.5 | 0.492 |
| MER-GCN[30](2020) | - | - | 42.71 | - | - | - |
| SLSTT[10](2021) | 73.17 | 0.724 | 75.806 | **0.753** | 72.388 | 0.640 |
| GEME[35](2021) | 64.63 | 0.6158 | 75.2 | 0.7354 | 55.88 | 0.4538 |
| Later[11](2022) | 73.17 | **0.7447** | 70.68 | 0.7106 | - | - |
| FDCN[38](2022) | - | - | 73.09 | 0.72 | 58.07 | 0.57 |
| KTGSL[51](2022) | 72.58 | 0.6820 | 75.64 | 0.6917 | - | - |
| Sparse Transformer[13](2022) | - | - | **76.11** | 0.7192 | **80.15** | **0.7547** |
| MSMMT(Ours) | **78.30** | 0.7216 | 71.31 | 0.6754 | 73.83 | 0.5881 |

proposed method in this paper performs the best on the SMIC database with an accuracy of 78.3%, which is 5.13% higher than the next best, and UF1 reaches 0.7216, which is at the next best level. The accuracy on the SAMM database reached 73.83%, which is second to the best result and at an advanced level. Although the Sparse transformer algorithm performs best on the SAMM dataset and the CASMEII dataset, the algorithm in this paper and the Sparse transformer algorithm have different ideas and each has its own advantages. The idea of the sparse transformer algorithm is to combine the CNN and transformer networks to extract spatiotemporal features, and sparsely integrate the weights of each encoder layer and input them to the next layer for further processing to obtain the key patches features. The algorithm in this paper, on the other hand, is designed based on a pure transformer network, which is designed to process the weights of all the previous layers before the last layer of input, integrating the weights of the previous multiple layers. As we mentioned earlier, effective attentional features are gradually accumulated and amplified in each layer, therefore, our algorithm is able to extract useful features that are attended to by multiple encoder layers. Compared to existing transformer-based algorithms, the algorithm in this paper focuses more on extracting features at different scales of the face, and is more capable of expressing features at different granularities, and can obtain comparable results even when using only apex frames.

*4.5.2 CDE*

Some similar and the latest micro-expression recognition algorithms are compared in Table 2, and it can be easily seen that our method achieves the best level with UF1 of 0.8160 and UAR of

Table2 Testing the micro-expression recognition results of the proposed MSMMT and the state-of-the-art method on the combined three datasets

|  | FULL | | SMIC-HS | | CASMEII | | SAMM | |
| --- | --- | --- | --- | --- | --- | --- | --- | --- |
|  | UF1 | UAR | UF1 | UAR | UF1 | UAR | UF1 | UAR |
| LBP-TOP[47*] | 0.5882 | 0.5785 | 0.2 | 0.528 | 0.7026 | 0.7429 | 0.3954 | 0.4102 |
| Bi-WOOF[47*] | 0.6296 | 0.6227 | 0.5727 | 0.5829 | 0.7805 | 0.8026 | 0.5211 | 0.5139 |
| OFF-ApexNet[53](2019) | 0.7196 | 0.7096 | 0.6817 | 0.6695 | 0.8764 | 0.8681 | 0.5409 | 0.5392 |
| STSTNet[22](2019) | 0.7353 | 0.7605 | 0.6801 | 0.7013 | 0.8382 | 0.8686 | 0.6588 | 0.681 |
| EMR[49](2019) | 0.7885 | 0.7824 | 0.7461 | 0.753 | 0.8293 | 0.8209 | **0.7754** | 0.7152 |
| STA-GCN[26](2021) | - | - | - | - | 0.7608 | 0.7096 | - | - |
| SLSTT[10](2021) | **0.816** | 0.790 | 0.724 | 0.707 | 0.901 | 0.885 | 0.715 | 0.642 |
| AUGCN[52](2021) | 0.7914 | 0.7933 | 0.7192 | 0.7215 | 0.8798 | 0.871 | 0.7715 | **0.7890** |
| MSMMT(Ours) | **0.8160** | **0.8191** | **0.7651** | **0.7780** | **0.9071** | **0.8878** | 0.7392 | 0.7163 |

0.8191 on the combined database. The performance on the SMIC and CASMEII datasets is outstanding, both outperforming the other algorithms, with UF1 reaching 0.7651 and UAR reaching 0.7780 for the SMIC database, UF1 reaching 0.9071 and UAR reaching 0.8878 for the CASMEII database. For the SAMM database, we can see that the results are also comparable but not at the optimal level. The above results show an important conclusion that our proposed method is effective and can obtain better representation of features in the CDE. The reason for being able to achieve this result, we believe that benefiting from multi-scale multi-modal, the network is able to learn the local features of the samples with sufficient samples than the single data set.

*4.5.3 Impact of $\alpha$*

This subsection discusses in detail the effect of the loss weight factor on the network, and we show the results for the three data sets from CDE's experiments as well as the combined set. As can be seen in Figure 8, the accuracy increases first and then decreases as the weighting factor $\alpha$ gradually increases. Although the parameters of $\alpha$ are different when the optimal results are obtained on the three datasets, comprehensive comparison shows that when $\alpha = 0.1$, the two modes have the best performance in the three datasets. Similarly, we selected the optimal results under the $\alpha$ parameter for the three data sets in the SDE experiment. The $\alpha$ parameter was different for the three datasets, with 0.1 for the SMIC and CASMEII database, 0.2 for the SAMM database.

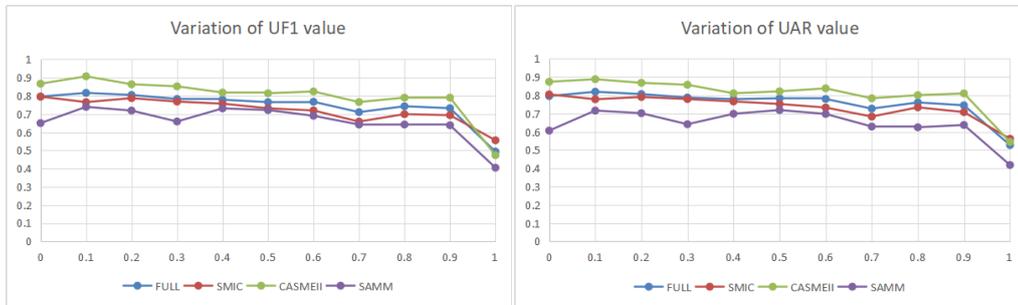

(a)                  (b)

Figure 8 Variation curves of UF1, UAR values with $\alpha$ for three categories (a) Variation of UF1 values (b) Variation of UAR values

## 5 Conclusion

In this work, we proposed a micro-expression recognition method based on multi-scale learning of bimodal features. The multi-scale features are learned by using ViT for dynamic imaging and optical flow features, and the patch features are weighted by using multi-headed self-attention weights in the network to obtain the most expressive facial region features and reduce the influence of irrelevant facial patches on the results. By combining cross-modal unsupervised contrastive learning, the information of texture and motion modal features are processed in the same category close to and different categories far from each other, enabling the network to fully use both features for expression learning. In this paper, a large number of experiments are conducted on three datasets, and the results obtained have good recognition rates, which fully demonstrate the effectiveness of the multi-scale algorithm proposed in this paper. In the future, we will further study the local feature expressions of micro-expressions on this basis and explore more meaningful features of different categories of micro-expressions.